\documentclass{article} 
\usepackage[final]{neurips_2022}

\usepackage{amsmath,amssymb,amsfonts}%
\usepackage{amsthm}%
\usepackage{mathrsfs}%
\usepackage{hyperref}
\usepackage{url}
\usepackage{siunitx}
\usepackage{multirow}

\usepackage{graphicx}
\usepackage[utf8]{inputenc} 
\usepackage[T1]{fontenc}    
\usepackage{hyperref}       
\usepackage{url}            
\usepackage{booktabs}       
\usepackage{amsfonts}       
\usepackage{nicefrac}       
\usepackage{microtype}      
\usepackage{xcolor}         

\usepackage{multirow}
\usepackage{tikz}
\usepackage{wrapfig}

\def\our{BayesianHMAML}
\def\ourb{BayesianHMAML-G}
\def\ourf{BayesianHMAML-CNF}

\def\L{\mathcal{L}}
\def\D{\mathcal{D}}
\def\E{\mathbb{E}}
\def\T{\mathcal{T}}
\def\S{\mathcal{S}}
\def\KL{\textit{KL}}

\usepackage{color}

\usepackage{algorithmicx}
\usepackage[ruled]{algorithm}
\usepackage{algpseudocode}
\usepackage{cancel}
\usepackage{makecell}

\title{Hypernetwork approach to Bayesian MAML}

\author{%
  P. Borycki, P. Kubacki,  M. Przewi\k{e}źlikowski, T. Ku\'smierczyk, J. Tabor, P. Spurek \\
  Faculty of Mathematics and Computer Science, \\
  Jagiellonian University, Kraków, Poland\\
  \texttt{przemyslaw.spurek@gmail.com} \\
}

\begin{document}

\maketitle


\begin{abstract}
   The main goal of Few-Shot learning algorithms is to enable learning from small amounts of data. One of the most popular and elegant Few-Shot learning approaches is Model-Agnostic Meta-Learning (MAML). The main idea behind this method is to learn the shared universal weights of a meta-model, which are then adapted for specific tasks. However, the method suffers from over-fitting and poorly quantifies uncertainty due to limited data size.
    Bayesian approaches could, in principle, alleviate these shortcomings by learning weight distributions in place of point-wise weights. Unfortunately, previous modifications of MAML are limited due to the simplicity of Gaussian posteriors, MAML-like gradient-based weight updates, or by the same structure enforced for universal and adapted weights.

    In this paper, we propose a novel framework for Bayesian MAML called \our{}, which employs Hypernetworks for weight updates. It learns the universal weights point-wise, but a probabilistic structure is added when adapted for specific tasks. 
    In such a framework, we can use simple Gaussian distributions or more complicated posteriors induced by Continuous Normalizing Flows.
\end{abstract}

\section{Introduction}

\begin{figure*}[htb] 
\begin{center} 
 \qquad MAML  \qquad \qquad BayesianMAML \qquad \qquad  \our{} \qquad \our{}\\ 
 \qquad \qquad  \qquad \qquad \qquad \qquad \qquad  \qquad \qquad  (Gaussian) \qquad \qquad \qquad   (CNF)\\  
\begin{tikzpicture}[scale=0.4]
    \node[inner sep=0pt] (russell) at (0,0)
    {\includegraphics[width=0.9\columnwidth]{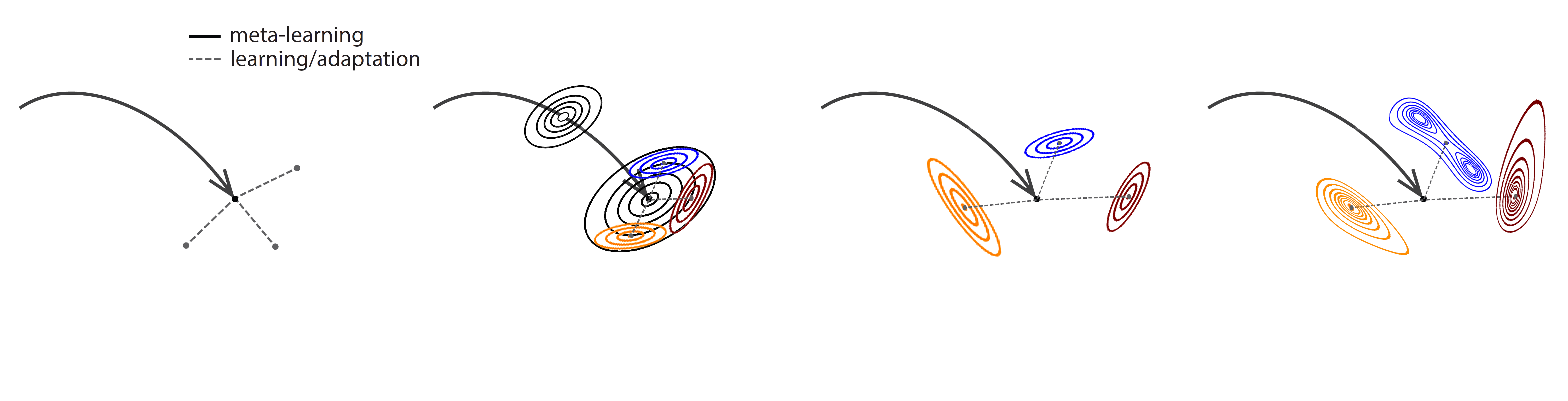} };
    \node[text width=2.5cm] at (-11.0, 2.4) { $\theta$}; 
    \node[text width=2.5cm] at (-11.0, -2.0) { $\theta'_i$};
    
    \node[text width=2.5cm] at (-3.0, 3.3) { $ \theta \sim \mathcal{N} (\mu, \sigma) $};         
    \node[text width=3.5cm] at (-4.5, -2.0) { $ \theta'_i \sim \mathcal{N} (\mu_{i}', \sigma_{i}'  ) | \mathcal{N} (\mu, \sigma  )$};
    
    \node[text width=2.5cm] at (7.0, 2.4) { $\theta$};   
    \node[text width=2.5cm] at (5.0, -2.0) { $ \theta'_i \sim \mathcal{N} (\mu_{i}', \sigma_{i}'  ) $};

    \node[text width=2.5cm] at (15.5, 2.4) { $\theta$};   
    \node[text width=3.5cm] at (15.0, -2.0) { $ \theta'_i \sim CNF_{\theta} $}; 

 \end{tikzpicture}

\end{center} 
  \vspace{-0.5cm}
  \caption{
  Comparison of four models: MAML~\citep{finn2017model}, BayesianMAML~\citep{yoon2018bayesian}, as well as \ourb{}, and \ourf{}. In the classic MAML, we have universal weights $\theta$, which are adapted to $\theta'_i$ for individual tasks $\mathcal{T}_i$. In BayesianMAML, the posterior distributions for individual small tasks are obtained in a few gradient-based updates from the universal distribution. In \ourb{}, we learn point-wise universal weights similar to MAML, but parameters of the specialized Gaussian posteriors are produced by a hypernetwork. Unlike BayesianMAML, the per-task distributions do not share a common prior distribution. In \ourf{}, the hypernetwork conditions a CNF, which can model arbitrary non-Gaussian posteriors.
  }   
\label{fig:teaser} 
\end{figure*}
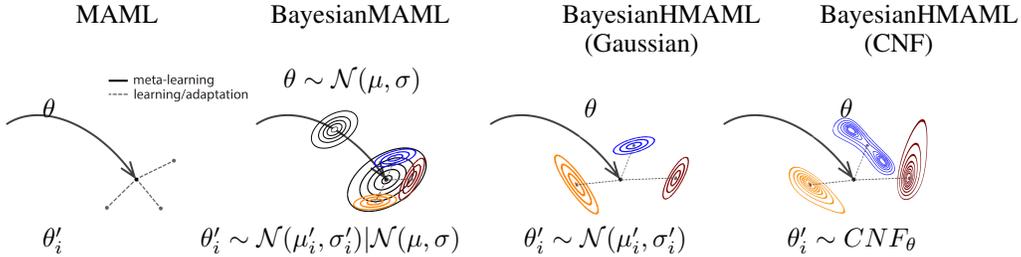

Few-Shot learning models easily adapt to previously unseen tasks based on a few labeled samples. One of the most popular and elegant among them is Model-Agnostic Meta-Learning (MAML)~\citep{finn2017model}. The main idea behind this method is to produce universal weights which can be rapidly updated to solve new small tasks (see the first plot in Fig.~\ref{fig:teaser}). However, limited data sets lead to two main problems. First, the method tends to overfit to training data, preventing us from using deep architectures with large numbers of weights. Second, it lacks good quantification of uncertainty, e.g., the model does not know how reliable its predictions are. Both problems can be addressed by employing Bayesian Neural Networks (BNNs)~\citep{mackay1992practical}, which learn distributions in place of point-wise estimates.

There exist a few Bayesian modifications of the classical MAML algorithm. Bayesian MAML~\citep{yoon2018bayesian}, Amortized bayesian meta-learning \citep{ravi2018amortized}, PACOH~\citep{rothfuss2021pacoh,rothfuss2020meta}, FO-MAML~\citep{nichol2018first}, MLAP-M~\citep{amit2018meta}, Meta-Mixture~\citep{jerfel2019reconciling} learn distributions for the common universal weights, which are then updated to per-task local weights distributions.  
The above modifications of MAML, similar to the original MAML, rely on gradient-based updates. Weights specialized for small tasks are obtained by taking a fixed number of gradient steps from the standard universal weights. Such a procedure needs two levels of Bayesian regularization and the universal distribution is usually employed as a prior for the per-task specializations (see the second plot in Fig.~\ref{fig:teaser}). However, the hierarchical structure complicates the optimization procedure and limits updates in the MAML procedure.

The paper presents \our{} -- a new framework for Bayesian Few-Shot learning. It simplifies the explained above weight-adapting procedure and thanks to the use of hypernetworks, enables learning more complicated posterior updates. Similar to the previous approaches, the final weight posteriors are obtained by updating from the universal weights. However, we avoid learning the aforementioned hierarchical structure by point-wise modeling of the universal weights. The probabilistic structure is added only later when specializing the model for a specific task. 
In \our{} updates from the universal weights to the per-task specialized ones are generated by hypernetworks instead of the previously used gradient-based optimization. Because hypernetworks can easily model more complex structures, they allow for better adaptations. In particular, we tested the standard Gaussian posteriors (see the third plot in Fig.~\ref{fig:teaser}) against more general posteriors induced by Continuous Normalizing Flows (CNF)~\citep{grathwohl2018ffjord} (see the right-most plot in Fig.~\ref{fig:teaser}).  

To the best of our knowledge, \our{} is the first approach that uses hypernetworks with Bayesian learning for Few-Shot learning tasks. 
Our contributions can be summarized as follows:
\begin{itemize}
    \item We introduce a novel framework for Bayesian Few-Shot learning, which simplifies updating procedure and allows using complicated posterior distributions. 
    \item Compared to the previous Bayesian modifications of MAML, \our{} employs the hypernetworks architecture for producing significantly more flexible weight updates. 
    \item We implement two versions of the model: \ourb{}, a classical Gaussian posterior and a generalized \ourf{}, relying on Conditional Normalizing Flows.
\end{itemize}

\section{Background}

This section introduces all the notions necessary for understanding our method.
We start by presenting the background and notation for Few-Shot learning. Then, we describe how the MAML algorithm works and introduce the general idea of Hypernetworks dedicated to MAML updates. Finally, we briefly explain Conditional Normalizing Flows.

\paragraph{The terminology} describing the Few-Shot learning setup is dispersive due to the colliding definitions used in the literature. 
Here, we use the nomenclature derived from the Meta-Learning literature, which is the most prevalent at the time of writing~\citep{wang2020generalizing,sendera2022hypershot}.

Let $\mathcal{S} = \{ (\mathbf{x}_l, \mathbf{y}_l) \}_{l=1}^L$ be a support-set containing $L$ input-output pairs with classes distributed uniformly. In the \textit{One-Shot} scenario, each class is represented by a single example, and $L=K$, where $K$ is the number of the considered classes in the given task. In the \textit{Few-Shot} scenarios, each class usually has from $2$ to $5$ representatives in the support set $\mathcal{S}$. 

Let $\mathcal{Q} = \{ (\mathbf{x}_m, \mathbf{y}_m) \}_{m=1}^M$ be a query set (sometimes referred to in the literature as a target set), with examples of $M$, where $M$ is typically an order of magnitude greater than $K$. Support and query sets are grouped in task $\T = \{\mathcal{S}, \mathcal{Q} \}$. Few-Shot models have randomly selected examples from the training set $\D = \{\T_n\}^N_{n=1}$ during training.
During inference, we consider task $\T_{*} = \{\mathcal{S}_{*}, \mathcal{X}_{*}\}$, where $\mathcal{S}_{*}$ is a set of support with known classes and $\mathcal{X}_{*}$ is a set of unlabeled query inputs. The goal is to predict the class labels for the query inputs $\mathbf{x} \in \mathcal{X}_*$, assuming support set $\mathcal{S}_{*}$ and using the model trained on the data $\D$.

\paragraph{Model-Agnostic Meta-Learning (MAML)} MAML~\citep{finn2017model} is one of the standard algorithms for Few-Shot learning, which learns the parameters of a model so that it can adapt to a new task in a few gradient steps. For the model, we use a neural network $f_{\theta}$ parameterized by weights $\theta$. Its architecture consists of a feature extractor (backbone) $E(\cdot)$ and a fully connected layer. The \emph{universal weights} $\theta= (\theta^{E}, \theta^{H})$ include $\theta^{E}$ for the feature extractor and $\theta^{H}$ for the classification head.

When adapting for a new task $\T_i =\{\mathcal{S}_i, \mathcal{Q}_i \} $,
the parameters $\theta$ are updated to $\theta'_i$.
Such an update is achieved in one or more gradient descent updates on $\T_i$. In the simplest case of one gradient update, the parameters are updated as follows:
$$
 \theta'_i = \theta - \alpha \nabla_{\theta} \L_{\T_i} (f_{\theta}), 
$$
where $\alpha$ is a step size hyperparameter. The loss function for a data set $\D$ is cross-entropy. The meta-optimization across tasks is performed via stochastic gradient descent (SGD):
$$
\theta \leftarrow  \theta - \beta \nabla_{\theta} \sum_{\T_i \sim p(T)} \L_{\T_i} (f_{\theta'_i})
$$
where $\beta$ is the meta step size (see Fig.~\ref{fig:teaser}).

\paragraph{Hypernetwork approche to MAML.} HyperMAML~\citep{przewikezlikowski2022hypermaml} is a generalization of the MAML algorithm, which uses non-gradient-based updates generated by hypernetworks~\citep{ha2016hypernetworks}. Analogically to MAML, it considers a model represented by a function $f_{\theta}$ with parameters $\theta$. When adapting to a new task $\T_i$, the parameters of the model $\theta$ become $\theta'_i$. 
Contrary to MAML, in HyperMAML the updated parameters $\theta'_i$ are computed using a hypernetwork $H_{\phi}$ as
$$
 \theta'_i = \theta + H_{\phi}( S_i, \theta ). 
$$
The hypernetwork $H_{\phi}$ is a neural network consisting of a feature extractor $E(\cdot)$, which transforms support sets into a lower-dimensional representation, and fully connected layers aggregate the representation.  To achieve permutation invariance, the embeddings are sorted according to their respective classes before aggregation. 

Similarly to MAML, the universal weights $\theta$ consist of the features extractor's weights $\theta_{E}$ and the classification head's weights $\theta_{H}$, i.e.,   $\theta= (\theta_{E}, \theta_{H})$.
However, HyperMAML keeps $\theta_{E}$ shared between tasks and updates only $\theta_{H}$, e.g.,
$$
 \theta'_i = (\theta'^{E}_i, \theta'^{H}_i) = (\theta^{E}_i, \theta^{H}_i + H_{\phi}( S_i, \theta )). 
$$ 

\begin{figure*}
\centering
      \includegraphics[width=\textwidth]{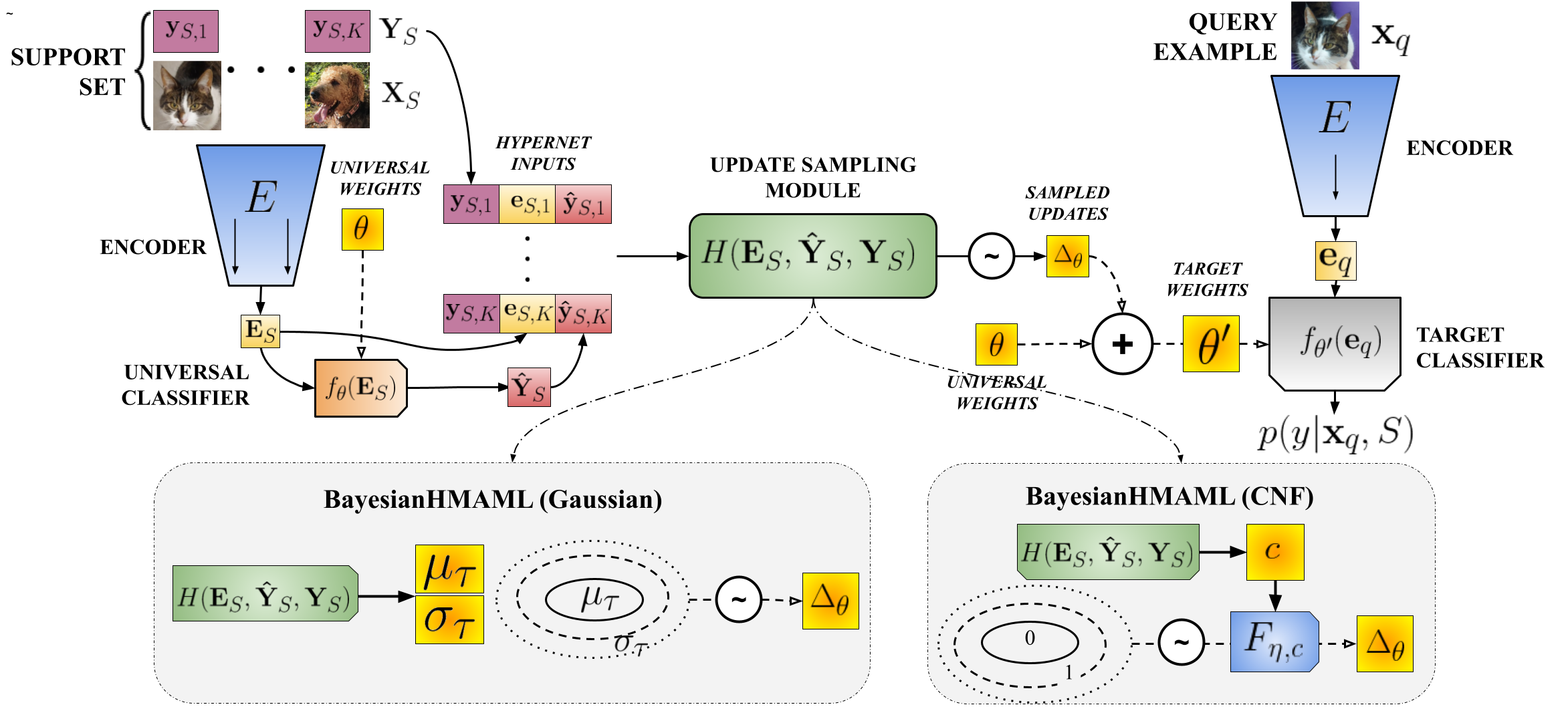}
  \caption{
  \our{}: Instead of updating the classifier weights with gradient descent, we use a hypernetwork $H$ to aggregate information from the support set $\S$ and to produce posterior parameters. First, the support set is transformed by an encoder. The embedded symbols $\mathbf{E}_{\mathcal{S}}$ are then concatenated with the original labels and predictions given by the universal weights. This representation is passed to a hypernetwork $H( \mathbf{E}_{\mathcal{S}},\mathbf{\hat{Y}}_{\mathcal{S}}, \mathbf{Y}_{\mathcal{S}})$  to produce posterior distributions. In our work, we consider two posterior variants: Gaussian and CNF-based. Using the hypernetwork for weight updates $\theta$ allows for larger and smarter adaptations of the posterior parameters.  In the end, we sample weight updates $\Delta \theta$ from the posterior distribution to obtain weights $\theta' = \theta + \Delta \theta$ dedicated to a specific task. 
  }
\label{fig:architecture} 
\end{figure*}

\paragraph{Continuous Normalizing Flows (CNF).}

The idea of normalizing flows \citep{dinh2014nice} relies on the transformation of a simple prior probability distribution $P_Z$ (usually a Gaussian one) defined in the latent space $Z$ into a complex one in the output space $Y$ through a series of invertible mappings: 
$F_\eta = F_K \circ \ldots \circ F_1: Z \to Y.$
The log-probability density of the output variable is given by the change of variables formula
$$
\log P_Y(y; \eta)   
 = \log P_Z(z) - \sum_{k=1}^K \log \left| \det \frac{\partial F_k}{\partial z_{k-1}} \right|,
$$
where $z = F_\eta^{-1}(y)$ and $P(y;\eta)$ denotes the probability density function induced by the normalizing flow with parameters $\eta$. The intermediate layers $F_i$ must be designed so that both the inverse map and the determinant of the Jacobian are computable.

The continuous normalizing flow \citep{chen2018neural} is a modification of the above approach, where instead of a discrete sequence of iterations, we allow the transformation to be defined by a solution to a differential equation
$ 
\frac{ \partial z(t)}{ \partial t} = g(z(t), t),
$
where $g$ is a neural network that has an unrestricted architecture. 
CNF, $F_{\eta}: Z \to Y$, is a solution of differential equations with the initial value problem $z(t_0) = y$, $\frac{\partial z(t)}{\partial t} =g_{\eta}(z(t), t)$. 
In such a case, we have
$$
\begin{array}{c}
 F_\eta(z) = F_{\eta}( z(t_0) ) =  z(t_0) + \int^{t_1}_{t_0} g_{\eta}(z(t), t) dt,\\[1em]
 F_{\eta}^{-1}(y) = y + \int_{t_1}^{t_0} g_{\eta}(z(t), t)dt,
\end{array}
$$
where $g_{\eta}$ defines the continuous-time dynamics of the flow $F_{\eta}$ and $z(t_1) = y$.

The logarithmic probability of $y \in Y$ can be calculated by:
$$
\log P_Y(y; \theta) = \log P_Z( F_{\theta}^{-1}(y) ) - \int^{t_1}_{t_0} \mathrm{Tr} \left( \frac{\partial g_{\theta}}{\partial z(t)} \right) dt.
$$

The main advantage of normalizing flows (both discrete and continuous ones) is that they are not restricted to any predefined class of distributions, e.g. Gaussian densities. For example, flow models can reliably describe densities of high-dimensional image data \citep{kingma2018glow}.

\section{BayesianHMAML -- a new framework for Bayesian learning }

In this section, we present \our{} -- a Bayesian extension of the classical MAML.
The most straightforward Bayesian treatment for such a model is to pose priors for the model parameters and learn their posteriors.  In particular, for MAML one needs to learn posterior distributions for both $\theta$ and $\theta'$. This naturally hints towards a hierarchical Bayesian model: $\theta \rightarrow \theta'_i \rightarrow \T_i$, which was previously proposed in \citep{ravi2018amortized,chen2022bayesian}. 
Hence, variational inference along with reparametrization gradients (i.e. Bayes by backpropagation~\citep{blundell2015weight}) is typically used, and the following objective (evidence lower bound) is maximized with respect to variational parameters $\lambda_i$ and $\psi$:
$$
\begin{array}{c}
    \L_{\D} =  \\[1em]
    \E_{q(\theta|\psi)}\!\!  
      \underbrace{\left[ \sum_i^N \E_{q(\theta'_i|\lambda_i)} \left[ \log p(\T_i|\theta'_i) - \KL\left( q(\theta'_i|\lambda_i) \| p(\theta'_i|\theta) \right) \right]  \right] }_{\L_{\T_i}}   
     - \KL(q(\theta|\psi) \| p(\theta)) 
\end{array}
$$
where $q(\theta'_i|\lambda_i)$ and $q(\theta|\psi)$ are respectively per-task posterior approximation and approximate posterior for the universal weights. They are tied together by the prior $p(\theta'_i|\theta)$.

The above formulation poses some challenges. First of all, updates are limited by the posterior of universal weights. Furthermore, the same distribution is used for the posterior weights for universal and updated weights. Finally, the Gaussian distribution is used for both posteriors.

\paragraph{Learned objective.}
We propose an alternative approach that alleviates the problems of previous attempts at Bayesian MAML. Contrary to them, we do not learn distributions for universal parameters $\theta$, but instead learn them in a pointwise manner. Distributional posteriors we learn only for individual task-specialized $\theta'_i$, where we assume their independence. Furthermore, we remove the coupling prior between $\theta$ and $\theta'_i$, and finally, we propose a basic non-hierarchical prior $p(\theta'_i)$ instead. 

\our{}'s learning objective takes the following form:
$$
\begin{array}{c}
    \L^{our}_D =  
    \sum_i^N  \E_{q(\theta'_i|\lambda_i(\theta, \S_i))  } \big[ \log p(\T_i|\theta'_i) -  
    \gamma \cdot \KL\left( q(\theta'_i|\theta, \lambda_i(\theta, \S_i)) \| p(\theta'_i) \right) \big],
\end{array}    
$$
where we used the standard normal priors for the weights of the neural network $f$, i.e., $p(\theta'_i) = \mathcal{N}(\theta'_i| 0, \mathbb{I})$. The hyperparameter $\gamma$ allows controlling the impact of the priors and compensating for model misspecification. Overall, the proposed modifications enable better optima for the objective and simplify the optimization landscape helping convergence.

\paragraph{Treatment of parameters.}
\our{} is a generalization of HyperMAML. Here however the weight updates result in posterior distributions instead of point-wise weights. In particular, when adapting a function $f_{\theta}$ with parameters $\theta$ to a task $\mathcal{T}_i$ the updated model’s parameters
$$\theta'_i \sim q(\theta | \lambda_i(\theta,S_i)).$$ 
In \our{} the parameters $\lambda_i$ are modeled by a hypernetwork as $$\lambda_i(\theta,S_i) := H_{\phi}( \theta, S_i ).$$
The hypernetwork $H_{\phi}$ takes support set $S_i$ and universal weights $\theta$ and when combined with the universal weights $\theta$ produces the posterior distribution $q$. Thanks to the hypernetwork, we obtain unconstrained updates and can model arbitrary posterior distributions, potentially improving over all previous non-hypernetwork models. 
Furthermore, the hypernetwork $H$ has a fixed number of parameters, regardless of how many tasks $\T_i$ are used for training. The amortized learning scheme has twofold benefits: (1) faster training; (2) regularization of learned parameters through shared architecture and common weights $\phi$.

We implemented two variants of \our{}: one using the standard Gaussian posterior and a generalized one with a flow-based posterior distribution.

\paragraph{Gaussian version.} \ourb{} is a simple realization of \our{}. In this approach, the hypernetwork $ H_{\phi}$ returns the mean update and covariance matrix of a Gaussian posterior:
$$
(\mu_{\theta}(\S_i) , \sigma_{\theta}(\S_i)) := H_{\phi}( \S_i, \theta ). 
$$
Weights are then sampled from the induced posterior: 
$$
 \theta'_i \sim \mathcal{N}(  \theta+\mu_{\theta}(\S_i), \sigma_{\theta}(\S_i)),
$$ 
We apply here the mean-field assumption, but note the standard deviations $\sigma$ are not entirely independent. Due to the used amortization scheme, they are tied together and to the means $\mu$ by shared weights $\phi$ of the hypernetwork. Posterior means however are additionally explicitly dependent by the universal weights $\theta$. Like the classic MAML, any change of $\theta$ affects all the values of $\theta'_i$.

\paragraph{CNF version.}
\ourf{} is a generalization of \ourb{}, where we use conditional flows to produce weight posteriors for the specialized tasks. Similar to \ourb{}, we employ a hypernetwork to amortize updates of the target model parameters. However, in the above model the hypernetwork outputs parameters of a Gaussian distribution, whereas in \ourf{} it is responsible for conditioning a flow $F_{\eta, C(\theta, \S_i) } (\cdot) $(see Fig.~\ref{fig:architecture} for comparison):
$$
C(\theta,\S_i) := H_{\phi}( \S_i, \theta ), 
$$
The conditioning vector $C$ is added to each layer of the flow to parameterize function $g_{\theta, C}$, so in the end, the flow  $F_{\eta, C}$ depends on trainable parameters $\eta$ and conditioning parameters $C$. 
Then, the posterior for a task $\T_i$ is obtained by a two-stage process:
\begin{align*}
    & \Delta\theta'_i \sim F_{\eta, C(\theta, \S_i)} \\
    & \theta'_i = \theta + \Delta\theta'_i, 
\end{align*}
where the shape of the posterior distribution is determined by the flow $F$, but its position, similarly to \ourb{}, mainly by the universal weights. 
From the implementation point of view, sampling from the conditioned flow also happens in two stages. First, we sample some $z$ from a flow prior and then, push this $z$ through a chain of deterministic transformations to obtain the final sample. Formally,
$$
\Delta\theta'_i := F_{\eta, C(\theta, \S_i) } (z), \mbox{ where } z \sim \mathrm{N}(0,t \cdot \mathbb{I}),
$$
where $t$ is a hyperparameter, we used $t=0.1$.

\paragraph{Architecture of \our{}.} The goal here is to predict the class distribution $p(\mathbf{y}|{x}_q, \mathcal{S})$, given a single query example $\mathbf{x}_q$, and a set of support examples $\mathcal{S}$. The architecture of \our{}  is illustrated in Fig.~\ref{fig:architecture}. Following MAML, we consider a parametric function $f_{\theta}$, which models the discriminative distribution for the classes. In addition, our architecture consists of a trainable encoding network $E(\cdot)$, which transforms data into a low-dimensional representation. The predictions are then calculated following 
$$
p(\mathbf{y}|{x}_q,\theta')=f_{\theta'}(\mathbf{e}_q),
$$
where $\mathbf{e}_q$ is the query example $\mathbf{x}_q$ transformed using encoder $E(\cdot)$, and $\theta'$ come from the posterior distribution for a considered task (either in \ourb{} or \ourf{} variant). In contrast to the MAML gradient-based adaptations, we predict weights directly from the support set using a hypernetwork. The hypernetwork observes support examples with the corresponding true labels and decides how the global parameters $\theta$ should be adjusted for a considered task. The two possible variants: \ourb{} and \ourf{} are illustrated (denoted by gray squares) in Fig.~\ref{fig:architecture}.

In \our{}, parameters of the learned posterior distribution are obtained by the hypernetwork $H_{\phi}(\theta, \S)$. First, each of the inputs from support set $\S$ is transformed by Encoder $E(\cdot)$ to obtain low-dimensional matrix of embeddings 
$
\mathbf{E}_{\mathcal{S}}=[\mathbf{e}_{\mathcal{S},1}, \dots, \mathbf{e}_{\mathcal{S},K}]^{\mathrm{T}}.
$ 
Next, the corresponding class labels for support examples, $\mathbf{Y}_{\mathcal{S}}= [\mathbf{y}_{\mathcal{S},1}, \dots, \mathbf{y}_{\mathcal{S}, K}]^\mathrm{T}$ are concatenated to the corresponding embeddings stored in the matrix $\mathbf{E}_{\mathcal{S}}$. Furthermore, we calculate the predicted values for the examples in the support set using the general model as
$
f_{\theta}(\mathbf{E}_{\mathcal{S}})=\mathbf{\hat{Y}_{\mathcal{S}}},
$
and also concatenate them into $\mathbf{E}_{\mathcal{S}}$. The predictions of the global model $f_{\theta}$ help identify classification errors and correct them by weight adaptation.  

Finally, the transformed support $\mathbf{E}_{\mathcal{S}}$, together with true labels $\mathbf{Y}_{\mathcal{S}}$, and the corresponding predictions $\mathbf{\hat{Y}_{\mathcal{S}}}$ 
from the model with universal weights are passed as input to the hypernetwork as $H( \mathbf{E}_{\mathcal{S}},\mathbf{\hat{Y}}_{\mathcal{S}}, \mathbf{Y}_{\mathcal{S}})$, which then predicts the posterior controlling parameters. In our case, the hypernetwork consists of fully-connected layers with ReLU activations.

\paragraph{Implementation details.}
Practical learning of \our{} is performed with stochastic gradients calculated w.r.t to the universal weights $\theta$ (i.e., $\nabla_\theta \L^{our}_{\T}$), the hypernetwork weights $\phi$ (i.e., $\nabla_\phi \L^{our}_{\T}$), and in case of \ourf{} also w.r.t $\eta$ (i.e., $\nabla_\eta \L^{our}_{\T}$), all of which have fixed sizes, which do not depend on the number of tasks $N$. We approximate the learning objective using mini-batches as
$$
\begin{array}{cc}
\L^{our}_{\T} = 
 \sum\limits_{\T_i \sim p(\T)}  \Big[  \frac{1}{P} \sum_{\theta'_i \sim q ( \theta, \lambda_i(\theta, \S_i))}  
 \big[ \L_{\T_i}(f_{ \theta_i'})  
- \gamma \KL( q (\theta_i' | \theta, \lambda_i(\theta, \S_i) ) \| \mathcal{N} (\theta_i' | 0, \mathbb{I}) ) \big]  \Big],
\end{array}
$$
where in each iteration we sample some number of tasks from $p(\T)$ and then, for each task, we sample $P=5$ samples $\theta'_i$ from the posterior $q$. How exactly is the sampling (and reparametrization) performed, depends on whether we use \ourb{} or \ourf{}. 

For \ourb{} the KL-divergence can be calculated in a closed form. For \ourf{} we use a Monte-Carlo estimate of $P$ samples:
$$
\begin{array}{cc}
\KL( \cdot )
=  \frac{1}{P} \sum_{ z \sim \mathcal{N} (0, t \cdot \mathbb{I}) } 
\big( \log \mathcal{N}\left( F^{-1}_{\eta, C }(\Delta \theta'_i | 0, t\cdot \mathbb{I} \right)  
+ \log \det |J| - \log N(\theta'_i|0, \mathbb{I}) 
\big),
\end{array}
$$
where $\Delta \theta'_i \equiv F_{\eta, C }(z)$ and $J$ is the flow transition Jacobian. 
Contrary to non-amortized methods, \our{} is easy to maintain (and scales well) since we need to store only a fixed number of parameters $\{\theta, \psi\}$ (or in case of \ourf{}: $\{\theta, \psi, \eta\}$). Finally, for the hyperparameter $\gamma$, we apply an annealing scheme~\citep{bowman2016generating}: the parameter $\gamma$ grows from zero to a fixed constant during training. The final value $\gamma_{max}$  is a hyperparameter of the model.

\section{Related Work}


The problem of Meta-Learning and Few-Shot learning \citep{hospedales2020metalearning,schmidhuber1992fast,bengio1992optimization} is currently one of the most important topics in deep learning, with the abundance of methods emerging as a result. They can be roughly categorized into three groups: 
Model-based methods, Metric-based methods, Optimization-based methods. In all these groups, we can find methods that use Hypernetworks and Bayesian learning (but not both at the same time). We briefly review the approaches below. 

Model-based methods aim to adapt to novel tasks quickly by utilizing mechanisms such as memory \citep{ravi2016optimization,mishra2018simple,zhen2020learning}, Gaussian Processes \citep{rasmussen2003gaussian,patacchiola2020bayesian,wang2021learning,sendera2021non}, or generating fast weights based on the support set with set-to-set architectures \citep{qiao2017fewshot,bauer2017discriminative,han2018learning,zhmoginov2022hypertransformer}. Other approaches maintain a set of weight templates and, based on those, generate target weights quickly through gradient-based optimization  such as \citep{zhao2020meta}.
The fast weights approaches can be interpreted as using Hypernetworks \citep{ha2016hypernetworks} -- models which learn to generate the parameters of neural networks performing the designated tasks.

Metric-based methods learn a transformation to a feature space where the distance between examples from the same class is small. The earliest examples of such methods are Matching Networks \citep{vinyals2016matching} and Prototypical Networks \citep{snell2017prototypical}. Subsequent works show that metric-based approaches can be improved by techniques such as learnable metric functions \citep{sung2018learning}, conditioning the model on tasks \citep{oreshkin2018tadam} or predicting the parameters of the kernel function to be calculated between support and query data with Hypernetworks \citep{sendera2022hypershot}. In~\citep{rusu2018meta}, authors introduce a meta-learning technique that uses a generative parameter model to capture the diverse range of parameters useful for distribution over tasks.

\begin{table*}[ht!]
\centering
\caption{Classification accuracy for inference on ${CUB}$ and ${mini-ImageNet}$ data sets in the $1$-shot and $5$-shot settings. The highest results are in bold and the second-highest in italic. }
\label{tab:conv45shotcubminiimagenet}

{
 \scriptsize
\begin{tabular}{@{}l@{}c@{}cc@{}c@{}}
\toprule
& \multicolumn{2}{c}{{CUB}}  & \multicolumn{2}{c}{{mini-ImageNet}} \\ 
{Method} & 1-shot & 5-shot  & 1-shot & 5-shot \\
\midrule
{Feature Transfer} \citep{zhuang2020comprehensive}  & $46.19 \pm 0.64$ &  $68.40 \pm 0.79$ & $39.51 \pm 0.23$ & $60.51 \pm 0.55$ \\

{ProtoNet} \citep{snell2017prototypical} &  $52.52 \pm 1.90$ & $75.93 \pm 0.46$ & $44.19 \pm 1.30$ & $64.07 \pm 0.65$ \\
{MAML} \citep{finn2017model} & $56.11 \pm 0.69$ & $74.84 \pm 0.62$ &  $45.39 \pm 0.49$  & $61.58 \pm 0.53$ \\
{MAML++} \citep{antioniou2018howto}   & -- & -- & $52.15 \pm 0.26 $ & $\mathit{68.32 \pm 0.44 }$\\
{FEAT} \citep{han2018learning} & $\mathbf{68.87 \pm 0.22}$ & $\mathbf{82.90 \pm 0.15}$  &  $\mathbf{55.15 \pm 0.20}$ & $\mathbf{71.61 \pm 0.16}$ \\

{LLAMA} \citep{grant2018recasting} & --  & -- &$49.40 \pm 1.83$  & -- \\
{VERSA} \citep{gordon2018meta} & -- &  -- & $48.53 \pm 1.84$  & $67.37 \pm 0.86$ \\
{Amortized VI} \citep{gordon2018meta} & -- & -- & $44.13 \pm 1.78$  & $55.68 \pm 0.91$ \\
{DKT + BNCosSim} \citep{patacchiola2020bayesian} & $62.96 \pm 0.62$  & $77.76 \pm 0.62$  & $49.73 \pm 0.07$  & $64.00 \pm 0.09$ \\
{VAMPIRE} \citep{nguyen2020uncertainty}& -- & -- & $51.54 \pm 0.74$ &  $64.31 \pm 0.74$ \\
{ABML} \citep{ravi2018amortized} & $49.57 \pm 0.42$  & $68.94 \pm 0.16$ & $45.00 \pm 0.60$ & -- \\
{OVE PG GP + Cosine (ML)} \citep{snell2020bayesian}   & $63.98 \pm 0.43$ & $77.44 \pm 0.18$  & $50.02 \pm 0.35$ & $64.58 \pm 0.31$\\
{OVE PG GP + Cosine (PL)}  \citep{snell2020bayesian}  & $60.11 \pm 0.26$ & $79.07 \pm 0.05$  & $48.00 \pm 0.24$ & $67.14 \pm 0.23$ \\

\midrule

{Bayesian MAML} \citep{yoon2018bayesian}  &   $55.93 \pm 0.71$ & -- & $\mathit{53.80 \pm 1.46}$  & $64.23 \pm 0.69$  \\
{HyperMAML} \citep{przewikezlikowski2022hypermaml} & $66.11 \pm 0.28$ & $ 78.89 \pm 0.19 $ & $ 51.84 \pm 0.57 $  & $ 66.29 \pm 0.43 $ \\
\midrule
{\our{} (G)}   & $ 66.57 \pm 0.47 $ & $ 79.86 \pm 0.31 $ & $ 52.54 \pm 0.46 $  & $ 67.39 \pm 0.35 $ \\
{\our{}  (G)+adapt.} & $ \mathit{66.92 \pm 0.38 }$ & $ \mathit{80.47 \pm 0.38} $  & $ 52.69 \pm 0.38  $ & $ 68.24 \pm 0.47 $ \\

\midrule

{\our{} (CNF) } & $ 61.55 \pm 0.69 $ & $ 75.41 \pm 0.21 $ & $ 49.39 \pm 0.33 $  & $ 64.77 \pm 0.21 $ \\
{\our{} (CNF)+adapt.} & $ 62.15 \pm 0.51 $ & $ 75.69 \pm 0.32 $  & $ 49.61 \pm 0.24 $  & $ 65.48 \pm 0.43 $ \\

\bottomrule
\end{tabular}
}
\end{table*}


Optimization-based methods
such as MetaOptNet \citep{lee2019meta} is based on the idea of an optimization process over the support set within the Meta-Learning framework. Arguably, the most popular of this family of methods is Model-Agnostic Meta-Learning (MAML) \citep{finn2017model}. In literature, we have various techniques for stabilizing its training and improving performance, such as Multi-Step Loss Optimization  \citep{antioniou2018howto}, or using the Bayesian variant of MAML \citep{yoon2018bayesian}.

Due to a need for calculating second-order derivatives when computing the gradient of the meta-training loss, training the classical MAML introduces a significant computational overhead. The authors show that in practice, the second-order derivatives can be omitted at the cost of small gradient estimation error and minimally reduced accuracy of the model \citep{finn2017model,nichol2018first}. Methods such as iMAML and Sign-MAML propose to solve this issue with implicit gradients or Sign-SGD optimization \citep{rajeswaran2019meta,fan2021signmaml}.
The optimization process can also be improved by training the base initialization \citep{munkhdalai2017meta,rajasegaran2020pamela}.
Furthermore, gradient-based optimization for few-shot tasks can be discarded altogether in favor of updates generated by hypernetworks~\citep{przewikezlikowski2022hypermaml}.


Classical MAML-based algorithms have problems with over-fitting. To address this problem, we can use the Bayesian models~\citep{ravi2018amortized,yoon2018bayesian,grant2018recasting,jerfel2019reconciling,nguyen2020uncertainty}. 
In practice, the Bayesian model contains two levels of probability distribution on weights. We have Bayesian universal weights, which are updated for different tasks \citep{grant2018recasting}. Its leads to a hierarchical Bayes formulation. Bayesian networks perform better
in few-shot settings and reduce
over-fitting. Several variants of the hierarchical Bayes model have been proposed based on different Bayesian inference methods \citep{finn2018probabilistic,yoon2018bayesian,gordon2018meta,nguyen2020uncertainty}. 
Another branch of probabilistic methods is represented by PAC-Bayes based method \citep{chen2022bayesian,amit2018meta,rothfuss2021pacoh,rothfuss2020meta,ding2021bridging,farid2021generalization}. In the PAC-Bayes framework, we use the Gibbs error when sampling priors. But still, we have a double level of Bayesian networks. 

In~\citep{rusu2018meta}, authors introduce a meta-learning technique using a generative parameter model to capture the diverse range of parameters useful for task distribution.
In VERSA \citep{gordon2019meta}, authors use amortization networks to produce distribution over weights directly.

In the paper, we propose \ourb{}, which uses probability distribution update only for weight dedicated to small tasks. Thanks to such a solution, we produce significantly larger updates.



\section{Experiments}

In our experiments, we follow the unified procedure proposed by \citep{chen2019closer}.
We split the data sets into the standard train, validation, and test class subsets, used commonly in the literature \citep{ravi2016optimization,chen2019closer,patacchiola2020bayesian}. We report the performance of both variants of \our{} averaged over three training runs for each setting. 

We report results for \our{} and for the model with adaptation. 
In the case of {\our{} + adaptation}, we tune a copy of the hypernetwork on the support set separately for each validation task. This way, we ensure that our model does not take unfair advantage of the validation tasks. In the case of hypernetwork-based approaches adaptation is a common strategy introduced by~\citep{sendera2022hypershot}.

First, we consider a classical Few-Shot learning scenario on two data sets: Caltech-USCD Birds ({CUB}) and {mini-ImageNet}. 

In the case of CUB, \ourb{} obtains the second-best score in the 1-shot and 5-shot settings. In the case of ${mini-ImageNet}$, we report comparable results comparable to other methods. We emphasize that \our{} obtains the best score in the area of Bayesian \citep{grant2018recasting,gordon2018meta,patacchiola2020bayesian,ravi2018amortized,snell2020bayesian,yoon2018bayesian} and MAML-based methods \citep{finn2017model,przewikezlikowski2022hypermaml}, save for MAML++ \citep{antioniou2018howto}.

\begin{table*}[t]
\caption{Classification accuracy for inference on cross-domain data sets ({Omniglot}$\rightarrow${EMNIST} and {mini-ImageNet}$\rightarrow${CUB}), in the $1$-shot and $5$-shot settings. The highest results are marked in \textbf{bold} and the second-highest in \textit{italics}. }
\centering
{
 \scriptsize
\begin{tabular}{@{}l@{}c@{}cc@{}c@{}}
\toprule
{} & \multicolumn{2}{c}{{Omniglot}$\rightarrow${EMNIST}} & \multicolumn{2}{c}{{mini-ImageNet}$\rightarrow${CUB}} \\
 {{Method}} & {1-shot}& {5-shot} & {1-shot} & {5-shot} \\ 
\midrule
 {{Feature Transfer}}  \citep{zhuang2020comprehensive} & 64.22 $\pm$  {1.24} & 86.10 $\pm$  {0.84} & 32.77 $\pm$  {0.35} & 50.34 $\pm$  {0.27}\\
 {{ProtoNet}} \citep{snell2017prototypical} & 72.04 $\pm$  {0.82} & 87.22 $\pm$  {1.01} & 33.27 $\pm$  {1.09} & 52.16 $\pm$  {0.17} \\
{{MAML}} \citep{finn2017model} & 74.81 $\pm$  {0.25} & 83.54 $\pm$  {1.79}  & 34.01 $\pm$  {1.25} &48.83 $\pm$  {0.62} \\

 {{DKT}} \citep{patacchiola2020bayesian} & 75.40 $\pm$  {1.10} &  $\mathbf{ 90.30 \pm  0.49}$  & $\mathbf{40.14 \pm  {0.18}}$ & $\mathit{56.40 \pm {1.34}}$ \\


{OVE PG GP + Cosine (ML)} \citep{snell2020bayesian} & $68.43 \pm 0.67 $ & $86.22 \pm 0.20 $ & $39.66 \pm 0.18$ & $55.71 \pm 0.31$ \\
{OVE PG GP + Cosine (PL)}  \citep{snell2020bayesian}  & $ 77.00 \pm 0.50 $ & $87.52 \pm 0.19 $ & $\mathit{37.49 \pm 0.11}$ & $\mathbf{57.23 \pm 0.31}$ \\
\midrule
{Bayesian MAML} \citep{yoon2018bayesian} & $63.94 \pm 0.47$ & $65.26 \pm 0.30 $ & $33.52 \pm 0.36 $ & $51.35 \pm 0.16$ \\
{HyperMAML} \citep{przewikezlikowski2022hypermaml} & $ 79.07 \pm 1.09 $ & $ {89.22 \pm 0.78}$ & $ 36.32 \pm 0.61 $ & $ 49.43 \pm 0.14 $ \\ 
\midrule 
{\our{} (G)} & $ \mathit{80.95 \pm 0.46} $ & $ 89.21 \pm 0.27 $ & $ 36.90 \pm 0.34 $  & $ 49.24 \pm 0.38 $ \\
{\our{} (G)+adapt.} & $\mathbf{81.05 \pm 0.47} $ & $\mathit{ 89.76 \pm 0.26 }$  & $ 37.23 \pm 0.44 $  & $ 50.79 \pm 0.59 $  \\
\midrule
{\our{} (CNF)} & $ 72.02 \pm 0.56 $ & $ 82.36 \pm 0.12 $ & $ 33.77 \pm 0.30 $  & $ 44.09 \pm 0.32 $ \\
{\our{} (CNF)+adapt.} & $ 72.54 \pm 0.36 $ & $ 82.63 \pm 0.37 $  & $ 34.67 \pm 0.35 $  & $ 45.14 \pm 0.27$  \\
\bottomrule
\hline
\end{tabular}
}
\label{tab:crossdomain_accuracy}
\end{table*}

In the cross-domain adaptation setting, the model is evaluated on tasks from a different distribution than the one on which it had been trained. 
We report the results in Table \ref{tab:crossdomain_accuracy}. In the task of 1-shot {Omniglot}$\rightarrow${EMNIST} classification, \ourb{} achieves the best result. The $5$-shot  {Omniglot}$\rightarrow${EMNIST} classification task \ourb{} yields comparable results to baseline methods. In the {mini-ImageNet}$\rightarrow${CUB} classification, our method performs comparably to baseline methods such as MAML and ProtoNet.


It needs to be highlighted that in all experiments \ourb{} achieves better performance than \ourf{}. 
It is caused mainly by the fact that the Gaussian posterior, whenever needed, can easily degenerate to a near-point distribution. In the case of the Flow-based model, the weight distributions are more complex and learning is significantly harder. 
 
\begin{figure}[htb] 
\begin{center} 
 \includegraphics[width=0.45\textwidth]{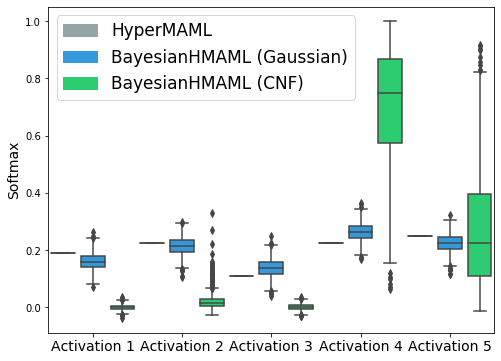}
\end{center} 
  \caption{
  Predictions of HyperMAML, \ourb{}, and \ourf{} on out-of-distribution data. 
  We trained the models on cross-domain data Omniglot$\rightarrow$EMNIST in a 5-shot setting, and then, for each of them, we sampled one thousand predictions. We show results on activations of 5 classes. Note, Bayesian models exhibit high uncertainty on out-of-distribution data.
  Furthermore, \ourf{} produces the most diverse predictions.
  } 
\label{fig:uncertainty} 
\end{figure}

The primary reason for using Bayesian approaches is better uncertainty quantification. Our models always give predictions for elements from support and query sets similar to the ones by HyperMAML and MAML. What is however crucial, we observe higher uncertainty in the case of elements from out of distribution. To illustrate that we trained \our{} on cross-domain adaptation setting {Omniglot}$\rightarrow${EMNIST}. Then, we sampled testing tasks from {EMNIST} during the evaluation and we sampled one thousand different weights from the distribution for our support set. Results are shown in Fig.~\ref{fig:uncertainty}.

\section{Conclusions}
\label{sec:conclusion}
In this work, we introduced \our{} -- a novel Bayesian Meta-Learning algorithm strongly motivated by MAML. 
In \our{}, we have universal weights trained in a point-wise manner, similar to MAML, and Bayesian updates modeled with hypernetworks. Such an approach allows for significantly larger updates in the adaptation phase and better uncertainty quantification.
Our experiments show that \our{}  outperforms all Bayesian and MAML-based methods in several standard Few-Shot learning benchmarks and in most cases, achieves results better or comparable to other state-of-the-art methods. Crucially, \our{} can be used to estimate the uncertainty of the predictions, enabling possible applications in critical areas of deep learning, such as medical diagnosis or autonomous driving.



\section{Appendix: Training details}

In this section, we present details of the training and architecture overview.

\subsection{Architecture details}

\paragraph{Encoder}

For each experiment described in the main body of this work, we utilize a shallow convolutional encoder (feature extractor), commonly used in the literature \citep{finn2017model,chen2019closer,patacchiola2020bayesian}.  This encoder consists of four convolutional layers, each consisting of a convolution, batch normalization, and ReLU nonlinearity. Each convolutional layer has an input and output size of 64, except for the first layer, where the input size equals the number of image channels. We also apply max-pooling between each convolution, decreasing the resolution of the processed feature maps by half. The output of the encoder is flattened to process it in the next layers. 

In the case of the {Omniglot} and {EMNIST} images, such encoder compresses the images into 64-element embedding vectors, which serve as input to the Hypernetwork (with the above-described enhancements) and the classifier. However, in the case of substantially larger {mini-ImageNet} and {CUB} images, the backbone outputs a feature map of shape $[64 \times 5 \times 5]$, which would translate to $1600$-element embeddings and lead to an over parametrization of the Hypernetwork and classifier which processes them and increase the computational load. Therefore, we apply an average pooling operation to the obtained feature maps and ultimately also obtain embeddings of shape $64$. Thus, we can use significantly smaller Hypernetworks.

\paragraph{Hypernetwork} 

The Hypernetwork transforms the enhanced embeddings of the support examples of each class in a task into the updates for the portion of classifier weights predicting that class. It consists of three fully-connected layers with ReLU activation function between each consecutive pair of layers. In the hypernetwork, we use a hidden size of $256$ or $512$.

\paragraph{Classifier} The universal classifier is a single fully-connected layer with the input size equal to the encoder embedding size (in our case 64) and the output size equal to the number of classes. When using the strategy with embeddings enhancement, we freeze the classifier to get only the information about the behavior of the classifier. This means we do not calculate the gradient for the classifier in this step of the forward pass. Instead, gradient calculation for the classifier takes place during the classification of the query data.

\subsection{Training details}

In all of the experiments described in the main body of this work, we utilize the switch and the embedding enhancement mechanisms. We use the Adam optimizer and a multi-step
learning rate schedule with the decay of $0.3$ and learning rate starting from $0.01$ or $0.001$. We train \our{} for 4000 epochs on all the data sets, save for the simpler {Omniglot} $\rightarrow$ {EMNIST} classification task, where we train for 2048 epochs instead.

\subsection{Hyperparameters} 
\label{app:sec:hyperparameters}
Below, we outline the hyperparameters of architecture and training procedures used in each experiment.

\begin{table*}[h!]
    \scriptsize
    \centering
    \begin{tabular}{@{}lcccc@{}}
\toprule
 {hyperparameter} & {CUB} & {mini-ImageNet} & {mini-ImageNet $\rightarrow$ CUB} & {Omniglot $\rightarrow$ EMNIST} \\
\midrule
      learning rate & $0.01$ & $0.001$ & $0.001$ & $0.01$ \\
      Hyper Network depth & $3$ & $3$ & $3$ & $3$ \\
      Hyper Network width & $512$ & $256$ & $256$ & $512$ \\
      epochs no. & $4000$ & $4000$ & $4000$ & $2048$ \\
       milestones & $51, 550$ & $101, 1100$  & $101, 1100$ & $51, 550$ \\
      $\gamma$  & $1e-4$ & $1e-4$ & $1e-5$ & $0.001$ \\
      num. of samples (train) & $5$ & $7$ & $5$ & $5$ \\
\bottomrule
\end{tabular}
    \caption{Hyperparameters for each of conducted 1-shot experiments. (G)}
    \label{tab:efficiency1}
\end{table*}

\begin{table*}[h!]
    \scriptsize
    \centering
    \begin{tabular}{@{}l@{}cccc@{}}
\toprule
 {hyperparameter} & {CUB} & {mini-ImageNet} & {mini-ImageNet $\rightarrow$ CUB} & {Omniglot $\rightarrow$ EMNIST} \\
\midrule
      learning rate & $0.001$ & $0.001$ & $0.001$ & $0.01$ \\
      Hyper Network depth & $3$ & $3$ & $3$ & $3$ \\
      Hyper Network width & $256$ & $256$ & $256$ & $512$ \\
      epochs no. & $4000$ & $4000$ & $4000$ & $2048$ \\
      milestones & $101, 1100$ & $101, 1100$ & $101, 1100$ & $51, 550$ \\
      $\gamma$ & $1e-5$ & $1e-5$ & $1e-4$ & $0.001$ \\
      num. of samples (train) & $5$ & $5$ & $5$ & $5$ \\
\bottomrule
\end{tabular}

    \caption{Hyperparameters for each of the conducted 5-shot experiments. (G)}
    \label{tab:efficiency2}
\end{table*}

\begin{table*}[h!]
    \scriptsize
    \centering
    \begin{tabular}{@{}lcccc@{}}
\toprule
 {hyperparameter} & {CUB} & {mini-ImageNet} & {mini-ImageNet $\rightarrow$ CUB} & {Omniglot $\rightarrow$ EMNIST} \\
\midrule
      learning rate & $0.001$ & $0.001$ & $0.001$ & $0.001$ \\
      Hyper Network depth & $3$ & $3$ & $3$ & $3$ \\
      Hyper Network width & $512$ & $256$ & $256$ & $512$ \\
      epochs no. & $4000$ & $4000$ & $4000$ & $2048$ \\
       milestones & $51, 550$ & $101, 1100$  & $101, 1100$ & $51, 550$ \\
      $\gamma$ & $1e-6$ & $1e-6$ & $1e-6$ & $1e-4$ \\    
      num. of samples (train) & $5$ & $5$ & $5$ & $5$ \\
\bottomrule
\end{tabular}
    \caption{Hyperparameters for each of conducted 1-shot experiments. (CNF)}
    \label{tab:efficiency3}
\end{table*}

\begin{table*}[h!]
    \scriptsize
    \centering
    \begin{tabular}{@{}lcccc@{}}
\toprule
 {hyperparameter} & {CUB} & {mini-ImageNet} & {mini-ImageNet $\rightarrow$ CUB} & {Omniglot $\rightarrow$ EMNIST} \\
\midrule
      learning rate & $0.001$ & $0.001$ & $0.001$ & $0.001$ \\
      Hyper Network depth & $3$ & $3$ & $3$ & $3$ \\
      Hyper Network width & $256$ & $256$ & $256$ & $512$ \\
      epochs no. & $4000$ & $4000$ & $4000$ & $2048$ \\
      milestones & $101, 1100$ & $101, 1100$ & $101, 1100$ & $51, 550$ \\
      $\gamma$ & $1e-6$ & $1e-6$ & $1e-6$ & $1e-6$ \\
      num. of samples (train) & $5$ & $5$ & $5$ & $5$ \\
\bottomrule
\end{tabular}

    \caption{Hyperparameters for each of the conducted 5-shot experiments. (CNF)}
    \label{tab:efficiency4}
\end{table*}

\section{Appendix: Extended results}

We include an expanded version of Table 1 from the main manuscript in Table \ref{tab:app_conv45shotcubminiimagenet}, comparing our approach to a larger number of meta-learning methods.

\begin{table*}[ht!]
\centering
\caption{Classification accuracy for inference on ${CUB}$ and ${mini-ImageNet}$ data sets in the $1$-shot and $5$-shot settings. The highest results are in bold and the second-highest in italic. }
\label{tab:app_conv45shotcubminiimagenet}

{
 \scriptsize
\begin{tabular}{l@{\hspace*{5mm}}cccc}
\toprule
& \multicolumn{2}{c}{{CUB}}  & \multicolumn{2}{c}{{mini-ImageNet}} \\ 
{Method} & 1-shot & 5-shot  & 1-shot & 5-shot \\
\midrule
{ML-LSTM} \citep{ravi2016optimization} & --  & -- & $43.44 \pm 0.77$ & $60.60 \pm 0.71$ \\
{SNAIL} \citep{mishra2018simple} & -- & -- & $45.10$  & $55.20$ \\
{Feature Transfer} \citep{zhuang2020comprehensive}  & $46.19 \pm 0.64$ &  $68.40 \pm 0.79$ & $39.51 \pm 0.23$ & $60.51 \pm 0.55$ \\

{ProtoNet} \citep{snell2017prototypical} &  $52.52 \pm 1.90$ & $75.93 \pm 0.46$ & $44.19 \pm 1.30$ & $64.07 \pm 0.65$ \\
{MAML} \citep{finn2017model} & $56.11 \pm 0.69$ & $74.84 \pm 0.62$ &  $45.39 \pm 0.49$  & $61.58 \pm 0.53$ \\
{MAML++} \citep{antioniou2018howto}   & -- & -- & $52.15 \pm 0.26 $ & ${68.32 \pm 0.44 }$\\
{FEAT} \citep{han2018learning} & $\mathbf{68.87 \pm 0.22}$ & $\mathbf{82.90 \pm 0.15}$  &  $\mathbf{55.15 \pm 0.20}$ & $\mathbf{71.61 \pm 0.16}$ \\

{LLAMA} \citep{grant2018recasting} & --  & -- &$49.40 \pm 1.83$  & -- \\
{VERSA} \citep{gordon2018meta} & -- &  -- & $48.53 \pm 1.84$  & $67.37 \pm 0.86$ \\
{Amortized VI} \citep{gordon2018meta} & -- & -- & $44.13 \pm 1.78$  & $55.68 \pm 0.91$ \\
{Meta-Mixture} \citep{jerfel2019reconciling} & -- & -- & $49.60 \pm 1.50$  & $64.60 \pm 0.92$ \\
{Baseline++} \citep{chen2019closer} & $61.75 \pm 0.95$  & $78.51 \pm 0.59$ & $47.15 \pm 0.49$ & $66.18 \pm 0.18$ \\
{MatchingNet} \citep{vinyals2016matching} & $60.19 \pm 1.02$ & $75.11 \pm 0.35$ & $48.25 \pm 0.65$  & $62.71 \pm 0.44$ \\
{RelationNet} \citep{sung2018learning} & $62.52 \pm 0.34$  & $78.22 \pm 0.07$  & $48.76 \pm 0.17$ & $64.20 \pm 0.28$ \\
{DKT + CosSim} \citep{patacchiola2020bayesian} & $63.37 \pm 0.19$  & $77.73 \pm 0.26$ & $48.64 \pm 0.45$  & $62.85 \pm 0.37$ \\
{DKT + BNCosSim} \citep{patacchiola2020bayesian} & $62.96 \pm 0.62$  & $77.76 \pm 0.62$  & $49.73 \pm 0.07$  & $64.00 \pm 0.09$ \\
{VAMPIRE} \citep{nguyen2020uncertainty}& -- & -- & $51.54 \pm 0.74$ &  $64.31 \pm 0.74$ \\
{PLATIPUS} \citep{finn2018probabilistic} & -- & -- & $50.13 \pm 1.86$  & -- \\
{ABML} \citep{ravi2018amortized} & $49.57 \pm 0.42$  & $68.94 \pm 0.16$ & $45.00 \pm 0.60$ & -- \\
{OVE PG GP + Cosine (ML)} \citep{snell2020bayesian}   & $63.98 \pm 0.43$ & $77.44 \pm 0.18$  & $50.02 \pm 0.35$ & $64.58 \pm 0.31$\\
{OVE PG GP + Cosine (PL)}  \citep{snell2020bayesian}  & $60.11 \pm 0.26$ & $79.07 \pm 0.05$  & $48.00 \pm 0.24$ & $67.14 \pm 0.23$ \\
{FO-MAML} \citep{nichol2018first} & -- & -- & $48.70 \pm 1.84$ & $63.11 \pm 0.92$  \\
{Reptile} \citep{nichol2018first} & -- & -- & $49.97 \pm 0.32$ & $65.99 \pm 0.58$  \\
{VSM} \citep{zhen2020learning} & --  & -- & $\mathit{54.73 \pm 1.60}$ &  $68.01 \pm 0.90$ \\
{PPA} \citep{qiao2017fewshot} & --  & -- & $\mathit{54.53 \pm 0.40}$ & -- \\
{HyperShot} \citep{sendera2022hypershot} & $65.27 \pm 0.24$  & $79.80 \pm 0.16$ & $52.42 \pm 0.46$  & $68.78 \pm 0.29$ \\
{HyperShot+ adaptation} \citep{sendera2022hypershot} & $66.13 \pm 0.26$& $80.07 \pm 0.22 $ & $53.18 \pm 0.45$  & $69.62 \pm 0.2$ \\

\midrule

{iMAML-HF} \citep{rajeswaran2019meta} & --  & -- & $49.30 \pm 1.88$  & --\\
{SignMAML} \citep{fan2021signmaml} & -- & -- & $42.90 \pm 1.50$ & $60.70 \pm 0.70$ \\
{Bayesian MAML} \citep{yoon2018bayesian}  &   $55.93 \pm 0.71$ & -- & ${53.80 \pm 1.46}$  & $64.23 \pm 0.69$  \\
{Unicorn-MAML} \citep{ye2021unicorn} & -- & -- &  $54.89$ & --\\
{Meta-SGD} \citep{li2017metasgd} & -- & -- & $50.47 \pm 1.87$ & $64.03 \pm 0.94$ \\
{MetaNet} \citep{munkhdalai2017meta} & -- & -- &  $49.21 \pm 0.96$  & -- \\
{PAMELA} \citep{rajasegaran2020pamela} & -- & -- & $53.50 \pm 0.89$ & $\mathit{70.51 \pm 0.67}$ \\
{HyperMAML} \citep{przewikezlikowski2022hypermaml} & $66.11 \pm 0.28$ & $ 78.89 \pm 0.19 $ & $ 51.84 \pm 0.57 $  & $ 66.29 \pm 0.43 $ \\
\midrule
{\our{} (G)}   & $ 66.57 \pm 0.47 $ & $ 79.86 \pm 0.31 $ & $ 52.54 \pm 0.46 $  & $ 67.39 \pm 0.35 $ \\
{\our{}  (G) + adaptation} & $ \mathit{66.92 \pm 0.38 }$ & $ \mathit{80.47 \pm 0.38} $  & $ 52.69 \pm 0.38  $ & $ 68.24 \pm 0.47 $ \\

\midrule

{\our{} (CNF) } & $ 61.55 \pm 0.69 $ & $ 75.41 \pm 0.21 $ & $ 49.39 \pm 0.33 $  & $ 64.77 \pm 0.21 $ \\
{\our{} (CNF) + adaptation} & $ 62.15 \pm 0.51 $ & $ 75.69 \pm 0.32 $  & $ 49.61 \pm 0.24 $  & $ 65.48 \pm 0.43 $ \\

\bottomrule
\end{tabular}
}
\end{table*}

\bibliographystyle{plain}

\end{document}